# IVSS: Integration of Color Feature Extraction Techniques for Intelligent Video Search Systems

Avinash N Bhute
Dept. of Computer Technology
Veermata Jijabai Technological Institute
Matunga, Mumbai, India
e-mail: anbhute@gmail.com

B. B. Meshram
Dept. of Computer Technology
Veermata Jijabai Technological Institute
Matunga, Mumbai, India
e-mail: bbmeshram@vjti.org.in

*Abstract*—As large amount of visual Information is available on web in form of images, graphics, animations and videos, so it is important in internet era to have an effective video search system. As there are number of video search engine (blinkx, Videosurf, Google, YouTube, etc.) which search for relevant videos based on user "keyword" or "term", But very less commercial video search engine are available which search videos based on visual image/clip/video. In this paper we are recommending a system that will search for relevant video using color feature of video in response of user Query.

## I. INTRODUCTION

As in internet era most difficult task is to retrieve the relevant information in response to a query. To help a user in this context various search system/engine are there in market with different features. In web search era 1.0 the main focus was on text retrieval using link analysis. It was totally read only era. There was no interaction in between the user and the search engine i.e. after obtaining search result user have no option to provide feedback regarding whether the result is relevant or not. In web search era 2.0 the focus was on retrieval of data based on relevance ranking as well as on social networking to read, write, edit and publish the result. Due to Proliferation of technology the current search era based on contextual search. Where rather than ranking of a page focus is on content based similarity to provide accurate result to user.

The CBVR (Content Based Video Retrieval) have received intensive attention in the literature of video information retrieval since this area was started couple of years ago, and consequently a broad range of techniques has been proposed. The algorithms used in these systems are commonly divided into four tasks:

- Segmentation
- Extraction
- Selection, and
- Classification

The segmentation task splits the video into number of chunks or shots. The extraction task transforms the content of video into various content features. Feature extraction is the process of generating features to be used in the selection and classification tasks. A feature is a characteristic that can capture a certain visual property of an image either globally for the whole image, or locally for objects or regions. Feature selection reduces the number of features provided to the classification task. Those features which are assisting in discrimination are selected and which are not selected is discarded. The selected features are used in the classification task [2]. The figure 1 shows the content based video search systems with four primitive tasks.

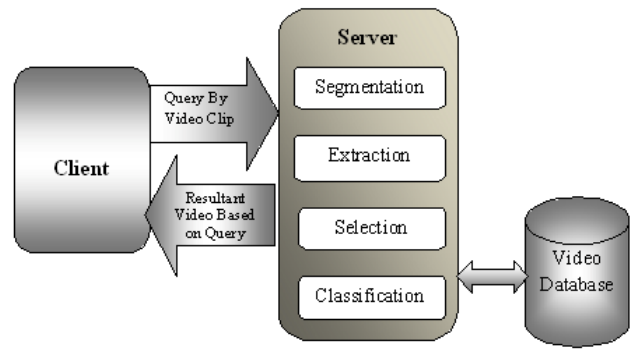

Fig. 1 Content based Video Search Systems and Its Task

Among these four activities feature extraction is critical because it directly influence the classification task. The set of features are the result of feature extraction. In past few years, the number of CBVR systems using different segmentation and extraction techniques, which proves reliable professional applications in Industry automation, social security, crime prevention, biometric security, CCTV surveillance [1], etc.

## II. PROPOSED CBVR SYSTEMS

Due to rapidity of digital information (Audio, Video) it become essential to develop a tool for efficient search of these media. With help of this paper we are proposing a Video Search system which will provide accurate and efficient result to a user query. The proposed system is a web based application as shown in fig.1 which consists of following processing:

*1) Client Side Processing:*
From client machine user can access the Graphical User Interface of the system. User can access and able to perform three tasks:
  a) Register the video
  b) Search the video
  c) Accept the efficient result from server.

*2. Server Side Processing:*
The core processing will be carried out at server side to minimize the overhead on client. Client will make a request for similar type of videos by providing query by video clip. On reception of this query by video clip, server will perform some processing on query video as well as on videos in its database and extract the video which are similar to query video. After retrieving the similar videos from the database, server will provide the list to the client in prioritized order. To accomplish this following operations are carried out at the server.

- Video segmentation and key frame extraction
- Feature Extraction
- Matching key frame features with feature database.
- Provide the resultant video to client in prioritized order.

*A. Video Segmentation and Key Frame Extraction*

The idea of segmenting an image into layers was introduced by Darrell and Pentland[3], and Wang and Adelson [5]. Darrel and Pentland [3] used a robust estimation method to iteratively estimate the number of layers and the pixel assignments to each layer. They show examples with range images and with optical flow. Wang and Adelson [5] is the seminal paper on segmenting video into layers. Affine model is fitted to blocks of optical flow, followed by a K-means clustering of these affine parameters. This step involves splitting and merging of layers, and therefore is not fixed. After the first iteration, the shape of regions is not confined to aggregate of blocks but is taken to a pixel level within the blocks. The results presented are convincing, though the edges of segments are not very accurate, most likely due to the errors in the computation of optical flow at occlusion boundaries Bergen et. al. [7] presents a method for motion segmentation by computing first the global parametric motion of the entire frame, and then finding the segments that do not fit the global motion model well. Irani and Peleg[6] incorporate temporal integration in this loop. A weighted aggregate of a number of frames is used to register the current image with the previous one. The object that is currently being compensated for thus becomes sharply into focus and everything else blurs out, improving the stability of the solution.

In order to extract valid information from video, process video data efficiently, and reduce the transfer stress of network, more and more attention is being paid to the video processing technology. The amount of data in video processing is significantly reduced by using video segmentation and key-frame extraction. Fig. 2 shows the basic framework of key frame extraction from a video .The explanation of each term is as given below:

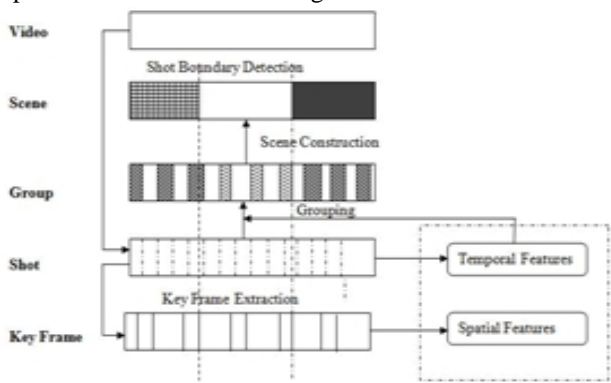

Fig. 2: The basic framework of the key frame extraction.

A shot is defined as the consecutive frames taken by a single camera without any significant change in the color content[8]. It shows a continuous action in an image sequence. Key frame is the frame which represents the salient visual contents of a shot [9]. The key frames extracted must summarize the characteristics of the video, and the image characteristics of a video can be tracked by all the key frames in time sequence. Depending on the complexity of the content of the shot, one or more frame can be extracted.

Video Scene is defines as collection of semantically related and temporally adjacent shots, depicting and conveying a high level concept or story. While shots are marked by physical boundaries, scenes are marked by semantic boundaries. Video group is an intermediate entity between the physical shots and semantic scenes and serves as the bridge between the two. For implementation of this phase we will use the Java Media Framework Plug in API's.

*B. Color Feature Extraction*

A key function in the content based video search system is feature extraction. A feature is a characteristic that can capture a certain visual property of an image either globally for the whole image, or locally for objects or regions. Color is an important feature for image representation which is widely used in image retrieval. This is due to the fact that color is invariance with respect to image scaling, translation and rotation [10].

Color is an important feature for image representation which is widely used in image retrieval. This is due to the fact that color is invariance with respect to image scaling, translation, and rotation [8]. The human eye is sensitive to colors, and color features are one of the most important elements enabling humans to recognize images [9]. Color features are, therefore, fundamental characteristics of the content of images. The algorithms which we are used to extract the color features of key frame are Average RGB, GCH, LCH, Color Moments.

*i) Average RGB*

Colors are commonly defined in three-dimensional color spaces. The color space models [13] can be differentiated as hardware-oriented and user-oriented. The hardware-oriented color spaces, including RGB and CMY are based on the three-color stimuli theory. The user-oriented color spaces, including HLS, HCV, HSV and HSB are based on the three human percepts of colors, i.e., hue, saturation, and brightness [11]. The RGB color space (see Figure 3) is defined as a unit cube with red, green, and blue axes; hence, a color in an RGB color space is represented by a vector with three coordinates. When all three values are set to 0, the corresponding color is black. When all three values are set to 1, the corresponding color is white [12].

The color histograms are defined as a set of bins where each bin denotes the probability of pixels in the image being of a particular color. A color histogram H for a given image is defined as a vector:

H={ H[*0*],H[*1*], ………H[*i*]…………H[*N*]}

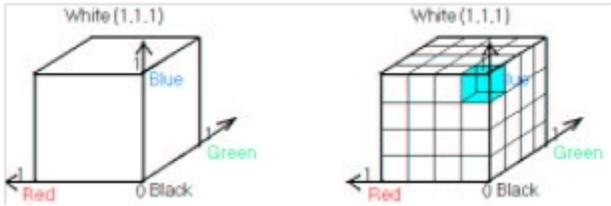

Fig. 3 RGB Color Space

where *i* represents a color in the color histogram and corresponds to a sub-cube in the RGB color space, $H[i]$ is the number of pixels in color *i* in that image, and N is the number of bins in the color histogram, i.e., the number of colors in the adopted color model.

*ii) Global Color Histogram*

As we have discussed, the color histogram depicts color distribution using a set of bins. Using the Global Color Histogram (GCH)[14], an image will be encoded with its color histogram, and the distance between two images will be determined by the distance between their color histograms. The following example (see Figure 4) shows how a GCH works.

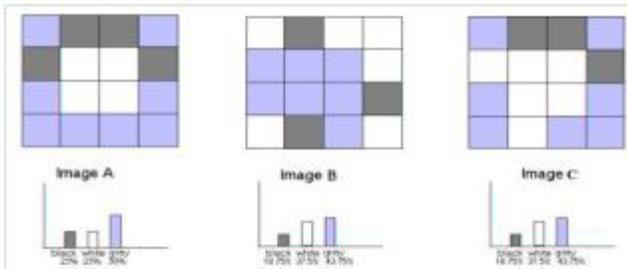

Fig. 4 Three images and their color histograms

In the sample color histograms there are three bins: black, white, and gray. We note the color histogram of image A:{25% ,25%,50%},the color histogram of image B:{18.75%, 37.5%, 43.75%} and image C has the same color histogram as image B. If we use the Euclidean distance metric to calculate the histogram distance, the distance between images A and B for GCH is:

$$d_{GCH(A,B)} = \sqrt{(a_1 - b_1)1^2 + (a_2 - b_2)^2 + (a_3 - b_3)1^2} \quad (1)$$

The distance between images A and C equals the distance between images A and B and the distance between images B and C is zero. The GCH is the traditional method for color-based image retrieval. However, it does not include information concerning the color distribution of the regions, so the distance between images sometimes cannot show the real difference between images.

*iii) Local Color Histogram*

This approach (referred to as LCH)[13][14] includes information concerning the color distribution of regions. The first step is to segment the image into blocks and then to obtain a color histogram for each block. An image will then be represented by these histograms. When comparing two images, we calculate the distance, using their histograms, between a region in one image and a region in same location in the other image. The distance between the two images will be determined by the sum of all these distances. If we use the square root of Euclidean distance as the distance between color histograms, the distance metric between two images Q and I used in the LCH will be defined as:

$$d_{LCH}(Q,I) = \sum_{K=1}^{M} \sqrt{\sum_{i=1}^{N}\left(H_Q^K[i] - H_I^K[i]\right)^2} \quad (2)$$

where M is the number of segmented regions in the images, N is the number of bins in the color histograms, and $H^k_Q[i]$ $H^K_I[i]$ is the value of bin *i* in color histogram $H^k_Q$ $H^K_I$ which represents the region k in the image Q(I).

The following examples use the same images A,B and C in Figure 5 to show how a LCH works and illustrate how we segment each image into 4 equally sized blocks. For the LCH, the distance between image A and B (see Figure 2.3) is calculated as follows:

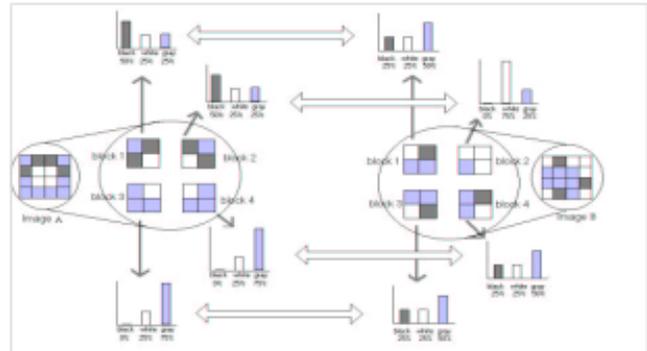

Fig.5 Using LCH to compute the distance between images A & B
$d_{LCH}(A,B)$ =1.768 & $d_{GCH}(A,B)$ =0.153

In some scenarios, using LCHs can obtain better retrieval effectiveness than using GCHs. The above examples show that the LCH overcomes the main disadvantage of the GCH, and the new distances between images may be more reasonable than those obtained using the GCH. However, since the LCH only compares regions in the same location, when the image is translated or rotated, it does not work well.

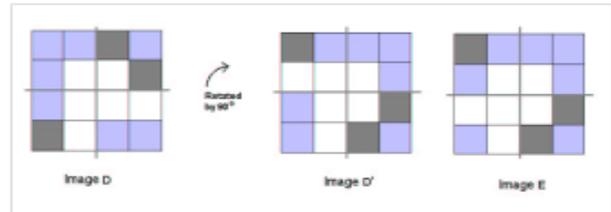

Fig. 6 An example showing that the LCH fails to compare images D and E

For example, in Figure 6 if image ( is rotated by $90^{o)}$, we get image D'. We can then see that image is very similar to image E with only two blocks being different. The distance between images D and E should equal to the distance between images D' and E. However, using LCH, the distance between images D and E will be greater than the distance between images D' and E. The reason for this discrepancy is that the LCH compares blocks only in the same location, but not necessarily in the proper locations. For example, using LCH, the north-west block in image D is compared with the north-west block in image E but, in fact, the north-west block in image D should be compared with the north-east block in image E. LCHs incorporate spatial information by dividing the image with fixed block segmentation. Some other approaches combine spatial information with color, using different partition schemes [15][16][17].

*iv) Color Moment*

Color moments[18] are measures that can be used to differentiate images based on their features of color. Once calculated, these moments provide a measurement for color similarity between images. These values of similarity can then be compared to the values of images indexed in a database for tasks like image retrieval. The basis of color moments lays in the assumption that the distribution of color in an image can be interpreted as a probability distribution. Probability distributions are characterized by a number of unique moments. These are mean, standard deviation and skewness. Moments are calculated for each of these channels in an image. An image therefore is characterized by 9 moments where 3 moments for 3 color channels, which are as follows.

Moment 1: Mean

$$E_i = \sum_N^{j=1} \frac{1}{N} P_{ij} \qquad (3)$$

Mean can be understood as the average color value in the i age.

Moment 2: Standard Deviation

$$\sigma_i = \sqrt{\left(\frac{1}{N}\sum_N^{j=1}(P_{ij} - E_i)^2\right)} \qquad (4)$$

The standard deviation is the square root of the variance of the distribution.

Moment 3: Skewness

$$s_i = \sqrt[3]{\left(\frac{1}{N}\sum_N^{j=1}(P_{ij} - E_i)^3\right)} \qquad (5)$$

Skewness can be understood as a measure of the degree of a symmetry in the distribution.

A function of the similarity between two image distributions is defined as the sum of the weighted differences between the moments of the two distributions. Formally this is:
$$D_{mom}(I_1, I_2) = \sum_{i=1}^{r} w_{i1}|E_i^1 - E_i^2| + w_{i2}|\sigma_i^1 - \sigma_i^2| + w_{i3}|S_i^1 - S_i^2| \qquad (6)$$

Where:
(H,I): are the two image distributions being compared

*v) Color Coherence Vector*

In Color Coherence Vector approach [19], each histogram bin is partitioned into two types, coherent and incoherent. If the pixel value belongs to a large uniformly-colored region then is referred to coherent otherwise it is called incoherent. In other words, coherent pixels are a part of a contiguous region in an image, while incoherent pixels are not. A color coherence vector represents this classification for each color in the image [20]. The coherence measure classifies pixels as either coherent or incoherent.

Coherent pixels are a part of some sizable contiguous region, while incoherent pixels are not. A color coherence vector represents this classification for each color in the image. Color coherence vector prevent coherent pixels in one image from matching incoherent pixels in another. This allows fine distinctions that cannot be made with color histograms.

*C. Integration Approach*

As Global color histogram method does not include information concerning the color distribution of the regions, Local Color Histogram fails when the image is translated or rotated, Color moment method do not encapsulate information about spatial correlation of colors. So to obtain the exact and accurate result from a video database we are using an integrated approach of color feature extraction methods where we are providing options to user to select any of combination of above described techniques. With this approach the feature vectors in different feature classes are combined into one overall feature vector. The system compares this overall feature vector of the query image to those of database images, using a predetermined similarity measurement. We are recommending here that to obtain a more relevant result user should select all the options provided in GUI.

The system is extracting the frames from a given video using Java Media Framework (JMF) Library. Input video supported by system are MPEG, AVI.

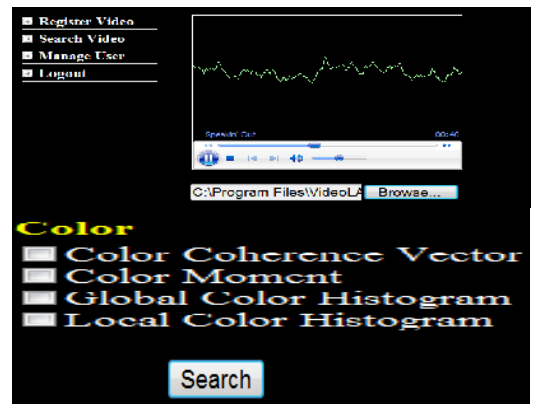

Fig. 7: GUI at client side for search

Kindly refer the Appendix A for results 1. Video to Frame Extraction and 2. Video to Key Frame Extraction

After extraction of each frame from inputted video, System is extracting the key frames by using the clustering based on Average RGB value.

Appendix A: Table 1 shows the every 5[th] frame of inputted video i.e car1.avi. Table 2 shows the every even frame of inputted video i.e car2.avi. Table 3 shows every even frame of inputted video i.e car3.avi Table 4 shows the result of key frame extraction of Car1.avi, Car2.avi and Car3.avi. Table 5 shows Query video with Key frames. Appendix A and B is Available online on following URL: http://www.vjti.ac.in/dept_comptech/Intenet_Multimedia/anb_comp/Appendix.asp

*D. Matching and Retrieval*

In retrieval stage of a Video search system, features of the given query video is also extracted. After that the similarity between the features of the query video and the stored feature vectors is determined. That means that computing the similarity between two videos can be transformed into the problem of computing the similarity between two feature vectors [21]. This similarity measure is used to give a distance between the query video and a candidate match from the feature data database as shown in Figure. 8.

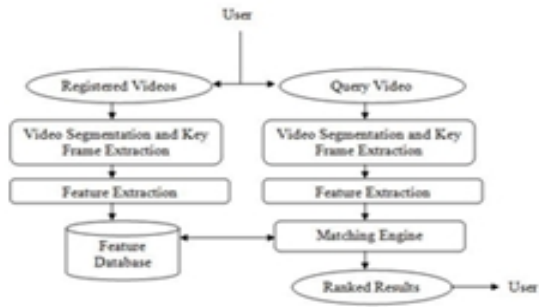

Fig. 8: Matching and retrieval of CBVR.

Kindly refer the Appendix B for color feature extraction Table 7 shows the behavior of color algorithms (CC: Color coherence, CM: color Moment, GC: Global color histogram, LC: Local color Histogram, AR: Average RGB) in response of query video (car4.avi).

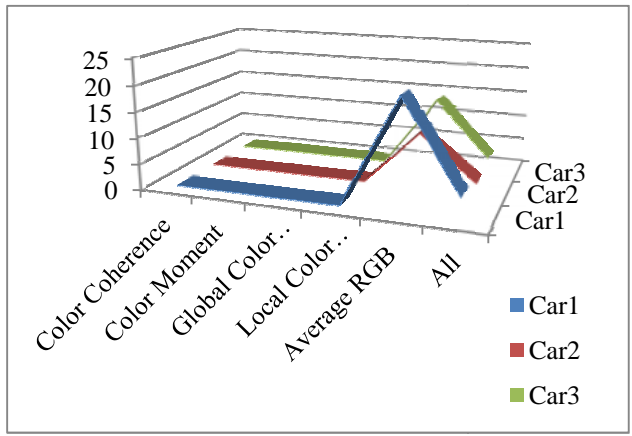

Fig. 9 Comparison of Color Feature Extraction Methods for query(car4.avi)

### III. CONCLUSION

The proposed systems facilitate the segmentation of the elementary shots in the long video sequence proficiently. Subsequently, the extraction of the color features using color histogram, color moment and color coherence vector is performed and the feature library is employed for storage purposes. One of advantage of this systems is user not only search for a relevant video but one can register his/her videos on the web server.